\documentclass{article} 
\usepackage{iclr2025_conference,times, custom}

\usepackage{amsmath,amsfonts,bm}

\def\eqref#1{equation~\ref{#1}}

\def\1{\bm{1}}

\DeclareMathAlphabet{\mathsfit}{\encodingdefault}{\sfdefault}{m}{sl}
\SetMathAlphabet{\mathsfit}{bold}{\encodingdefault}{\sfdefault}{bx}{n}

\usepackage{fancyhdr}
\usepackage{hyperref}
\usepackage{url}
\usepackage{booktabs}
\usepackage{array}
\usepackage[most]{tcolorbox}
\usepackage{mathtools}
\usepackage{subcaption}
\definecolor{tkcolor}{RGB}{224,223,255}
\newtcolorbox{takeaways}[1][]{
	width=\columnwidth,
	colback = tkcolor, 
	colframe = tkcolor, 
	boxsep=0pt,left=10pt,right=10pt,top=5pt,bottom=5pt,
	fontupper=\linespread{0.9}\selectfont,
	title=#1}

\title{When Modalities Conflict: How Unimodal Reasoning Uncertainty Governs Preference Dynamics in MLLMs}

\author{\textbf{Zhuoran Zhang}$^{1,2}$ \quad
\textbf{Tengyue Wang}$^{2,4}$ \quad
\textbf{Xilin Gong}$^{2,6}$ \quad
\textbf{Yang Shi}$^{1}$ \quad 
\textbf{Haotian Wang}$^{5}$ \\
\textbf{Di Wang}$^{2,3}$ \quad
\textbf{Lijie Hu}$^{7}$ \\
\\[0.5em]
\textsuperscript{1}Peking University \\
\textsuperscript{2}Provable Responsible AI and Data Analytics (PRADA) Lab \\
\textsuperscript{3}King Abdullah University of Science and Technology \\
\textsuperscript{4}South China University of Technology \\
\textsuperscript{5}Tsinghua University \\
\textsuperscript{6}University of Georgia  \\
\textsuperscript{7}MBZUAI
}

\iclrfinalcopy 
\begin{document}

\maketitle

\fancyhf{}                             
\renewcommand{\headrulewidth}{0.4pt}   
\renewcommand{\footrulewidth}{0pt}     
\fancyhead[L]{\itshape Preprint.} 
\fancyfoot[C]{\thepage}                
\pagestyle{fancy}
\thispagestyle{fancy}                  

\begin{abstract}
Multimodal large language models (MLLMs) must resolve conflicts when different modalities provide contradictory information, a process we term modality following. Prior work measured this behavior only with coarse dataset-level statistics, overlooking the influence of models’ confidence in unimodal reasoning. In this paper, we introduce a new framework that decomposes modality following into two fundamental factors: relative reasoning uncertainty ( the case-specific confidence gap between unimodal predictions) and inherent modality preference( a model’s stable bias when uncertainties are balanced).
To validate this framework, we construct a controllable dataset that systematically varies the reasoning difficulty of visual and textual inputs. Using entropy as a fine-grained uncertainty metric, we uncover a universal law: the probability of following a modality decreases monotonically as its relative uncertainty increases. At the relative difficulty level where the model tends to follow both modalities with comparable probability what we call the balance point, a practical indicator of the model’s inherent preference. Unlike traditional macro-level ratios, this measure offers a more principled and less confounded way to characterize modality bias, disentangling it from unimodal capabilities and dataset artifacts.
Further, by probing layer-wise predictions, we reveal the internal mechanism of oscillation: in ambiguous regions near the balance point, models vacillate between modalities across layers, explaining externally observed indecision. Together, these findings establish relative uncertainty and inherent preference as the two governing principles of modality following, offering both a quantitative framework and mechanistic insight into how MLLMs resolve conflicting information.
\end{abstract}
\section{Introduction}

Multimodal large language models (MLLMs) \citep{gpt4,gemini,qwen2vl,survey1,openai2024gpt4technicalreport} demonstrate powerful capabilities by processing information from various sources, like images and text, making them vital in applications ranging from web navigation \citep{openai_operator_2025} to aiding visually impaired users. However, a critical challenge arises when these modalities present conflicting information. For example, an image might show a blue car, while the accompanying text describes it as red. In such cases, the MLLM must resolve the conflict, leading to an observable behavior we term \textbf{modality following}: the model's final output aligns with the information from one modality over the other.

Prior studies \citep{zhang2025evaluatingsteeringmodalitypreferences,deng2025wordsvisionvisionlanguagemodels} have typically examined this phenomenon using coarse, dataset-level statistic: the ratio of text-following versus vision-following cases on a given set of conflicting inputs. This approach, however, often attempts to neutralize the model's unimodal capabilities by filtering for cases where the model can correctly answer based on either modality alone. This overlooks a crucial factor: the model's \emph{confidence} in each of its unimodal predictions. For the same instance, one model may produce the correct answer with high confidence while another does so with low confidence. 
Even within a single model, two different instances can elicit correct unimodal answers but with vastly different certainty levels. 
Such variations in underlying confidence directly influence the model’s final choice in multimodal settings and, consequently, shape the aggregate statistics of modality-following behavior.

To truly understand the modality-following  process, we propose that the static, dataset-level following statistics are emergent properties of two distinct underlying factors: (1) the \textbf{relative reasoning uncertainty} between the two modalities on a case-by-case basis, 
measured under unimodal inputs, which reflects the model's confidence gap between text-only and vision-only reasoning, and (2) a more stable, \textbf{inherent modality preference}, which we define as the model’s intrinsic leaning toward one modality when the reasoning uncertainties from both are perceived as equal. This leads to our central hypothesis:
\begin{quote}
\textbf{An MLLM's modality-following behavior is a dynamic process governed by the interplay between the relative reasoning uncertainty of the conflicting modalities and the model's own inherent preference.}
\end{quote}
In simpler terms, a model's decision to follow the text depends on whether the text's reasoning advantage (i.e., its low relative uncertainty compared to the image) is significant enough to overcome the model's potential inherent preference for vision.

We quantified the model's perceived uncertainty for each unimodal case using the \emph{output entropy} of its answer token, where a higher value indicates lower confidence \citep{shannon1948mathematical,farquhar2024detecting,zhang2024vluncertaintydetectinghallucinationlarge,cao2025amealignedmanifoldentropy}.Our overall analysis process is shown in Figure~\ref{fig:framework}. To validate the hypothesis, we constructed a controllable toy dataset that allows us to systematically and independently manipulate the reasoning difficulty of visual and textual inputs, thereby inducing varying levels of uncertainty in unimodal reasoning. The relationship between these two uncertainty scores was then used to define the relative uncertainty, forming the central axis for our analysis.

Our first goal was to verify if relative uncertainty indeed governs the model's final choice. By analyzing the model's outputs across our benchmark, we uncovered a clear and predictable pattern. As we systematically increased the reasoning uncertainty of one modality relative to the other, the model's probability of following that modality showed a consistent \textbf{monotonic decrease}. This finding confirms that modality following is not a fixed attribute but a fluid behavior that predictably shifts with the relative difficulty of unimodal inputs.

However, we observed that a model does not necessarily follow the modality with the lower relative uncertainty. Instead, each model possesses a unique threshold—a subjective \textbf{balance point} of uncertainty that it is willing to tolerate. This balance point reveals the model's \textbf{inherent preference}. For example, a model with a strong inherent preference for vision might only follow the text if the text is \emph{significantly} easier to process than the image.

Having established this behavioral relationship, we then sought to understand the internal mechanism behind it. \emph{Why} does a model hesitate or average its choices when the relative uncertainty is near its subjective balance point? To explore this, we categorized conflict scenarios into two types. In a \textbf{clear region}, where one modality is significantly less uncertain (i.e., much easier) than the other, the model quickly and stably commits to the easier modality in its early processing layers. In contrast, in the \textbf{ambiguous region} where both modalities have a similarly high or low level of uncertainty close to the model's balance point, the model will hesitate. This is visible internally as \textbf{``oscillations"}, where the model's top prediction repeatedly switches between the answer suggested by text and the one by vision across its layers. This internal oscillation provides a mechanistic explanation for the externally observed behavior of averaged following in uncertain situations. In summary, this paper makes three key contributions:
\begin{itemize}
    \item We propose a new framework that decomposes the observable ``modality following'' behavior into two core components: case-specific relative reasoning uncertainty and a model's stable inherent modality preference.
    \item Using a novel controllable dataset, we empirically discover a fundamental law: a model's probability of following a modality monotonically decreases as its relative reasoning uncertainty increases. We show how a model's inherent preference can be quantified as the balance point on this curve.
    \item We uncover the internal mechanism of ``oscillation'' within the model's layers, explaining why models hesitate and average their choices in ambiguous scenarios, thus linking internal dynamics to external behavior.
\end{itemize}
\section{Defining Conflicting Inputs and Quantifying Modality Following}
\label{sec:preliminary}
\begin{figure}[!t]
  \centering
\includegraphics[width=\linewidth]{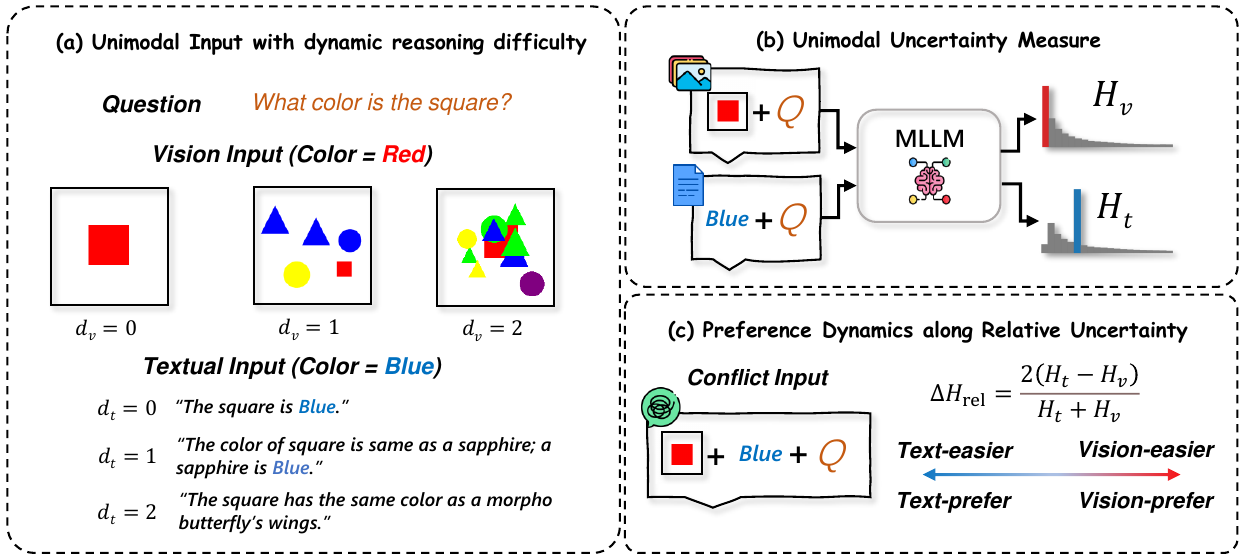}
    \caption{Overview of the analytical framework. \textbf{(a)} We create inputs with independently controllable visual ($d_v$) and textual ($d_t$) difficulty. \textbf{(b)} We measure the model's perceived uncertainty for each modality via output entropy ($H_v$, $H_t$). \textbf{(c)} We then use the relative uncertainty ($\Delta H_{rel}$) to analyze the model's choice when faced with a conflict.}
    \label{fig:framework}
\end{figure}

\paragraph{Conflicting Inputs.}
We define a \emph{conflicting input} as a triplet $(I, T, Q)$ consisting of an image $I$, a textual description $T$, and a question $Q$, 
such that the unimodal predictions of the MLLM $M_{\theta}$ disagree:
\[
Y_{v} = M_{\theta}(Q, I) \;\neq\; Y_{t} = M_{\theta}(Q, T).
\]
Here, $Y_{v}$ and $Y_{t}$ denote the predictions when the model relies solely on the visual or textual modality, respectively. 
For example in Figure~\ref{fig:framework}~(a), consider the question $Q=$ ``What is the color of the square?''.  
If the image $I$ shows a red square, while the text $T$ states ``The color of the square is the same as a morpho butterfly's wings'', 
then the image supports the answer ``red'' whereas the text suggests ``blue''.  
This forms a concrete instance of a conflicting input triplet $(I, T, Q)$. This setting requires the model to resolve contradictory cues and implicitly decide which modality to follow.

\paragraph{Macro-level Metrics for Modality Following.}
Given a conflicting input $x = (I, T, Q)$, the multimodal prediction is $Y_{m}=M_{\theta}(x)$. 
We categorize the outcome as \textbf{vision-following} if $Y_{m}=Y_{v}$, 
\textbf{text-following} if $Y_{m}=Y_{t}$, and \textbf{other} otherwise. 
To quantify the aggregate modality-following behavior on a dataset, we adopt the traditional approach of calculating following ratios. We define the text-following ratio (TFR) and vision-following ratio (VFR) as:
\[
\text{TFR} = 
\frac{|\{x : {Y}_{m}={Y}_{t}\}|}
     {|\{x : {Y}_{m}\in \{{Y}_{v}, {Y}_{t}\}\}|}, 
\quad 
\text{VFR} = 1 - \text{TFR}.
\]
These ratios offer a simple, macro-level statistic of a model's aggregate behavior. In subsequent sections, we will deconstruct how these statistics emerge from a deeper interplay between case-specific uncertainty and a model's inherent preference, which these ratios alone cannot capture.

\section{Preparing for the Analysis: A Controllable Dataset and an Uncertainty Metric}

To systematically investigate our central hypothesis: \textbf{that modality following is governed by relative uncertainty and inherent preference}, we must first establish a controlled experimental setup. This section details the two essential preparations for our analysis: (1) the construction of a novel dataset with independently controllable difficulty levels for both vision and text, and (2) the validation of entropy as the uncertainty metric, to precisely quantify the model's perceived reasoning difficulty in a fine-grained, modality-comparable manner.

\begin{figure}[htbp]
  \centering
\includegraphics[width=\linewidth]{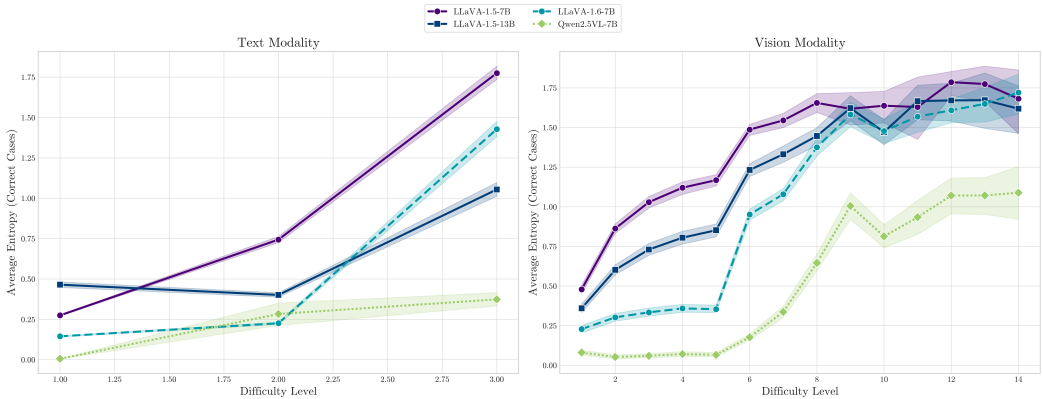}
    \caption{Unimodal Entropy Trends Across Difficulty Tiers. Average unimodal entropy for text (left) and vision (right) as a function of our designed difficulty tiers. Across all models, entropy consistently increases with difficulty, validating its use as a proxy for model-perceived uncertainty and revealing differences in model capabilities.}
    \label{fig:unimodal_entropy}
\end{figure}
\subsection{Constructing a Dataset with Controllable Difficulty}
\label{sec:methodology}

Existing benchmarks lack the ability to systematically vary the reasoning difficulty of each modality independently. To overcome this, we built a toy dataset where each multimodal instance is defined by a task type $\mathcal{T}$ and two integer-based \emph{design tiers}, $d_v$ and $d_t$, which control the complexity of the visual and textual inputs, respectively. 

We use the color recognition task as an example. As shown in Figure~\ref{fig:framework}(a),  the visual design tier ($d_v$) modulates perceptual difficulty by adding distractors, shrinking the target object, or introducing occlusions. A low $d_v$ might feature a single, clear red square, while a high $d_v$ might present it as a small, partially obscured object among many other colorful shapes. Similarly, the textual design tier ($d_t$) controls reasoning complexity. A low $d_t$ provides a direct (but conflicting) statement (e.g., ``The square is blue''), while a high $d_t$ requires multi-hop relational reasoning (e.g., ``The square shares its color with a morpho butterfly's wings''). We ensure that the conflicting color mentioned in text never appears among visual distractors, so each modality provides information independently.  By systematically pairing different levels of $d_v$ and $d_t$, we generate a structured landscape of conflict cases that spans a wide and predictable range of relative difficulty. Further details are in Appendix~\ref{app:dataset_stats}.

\subsection{Quantifying Perceived Uncertainty with Entropy}
\label{sec:entropy}

\paragraph{Entropy as proxy of perceived uncertainty.} While design tiers provide a human-interpretable notion of difficulty, our analysis requires a model-centric metric that reflects the model's \emph{own} perceived uncertainty. For this purpose, we employ the \textbf{Entropy} of the model's output distribution over the answer token \citep{shannon1948mathematical,cao2025amealignedmanifoldentropy}. Given a unimodal input $x$ (either vision-only or text-only), for example, consider a vision-only input where the question is ``What is the color of the square?'' and the image shows a red square. Its uncertainty is:
\begin{equation*}
H(x) = - \sum_{y \in \mathcal{V}} p(y \mid x) \, \log p(y \mid x),
\end{equation*}
where $\mathcal{V}$ is the token vocabulary. A low entropy value indicates a confident, sharp prediction (e.g., the probability for ``red'' is high, and near zero for other tokens), whereas a high entropy value would suggest that the model also considers alternative tokens (e.g., ``orange,'' ``brown''), 
revealing greater uncertainty about its own prediction. Since the output is always in the same token space, entropy serves as a unified and comparable measure of perceived uncertainty across both modalities, which we denote as $H^{(v)}$ for vision and $H^{(t)}$ for text.

\paragraph{Analysis of Unimodal Entropy Trends.} To validate that entropy reliably captures our designed difficulty, we measured it across different models and tiers, with the results presented in Figure~\ref{fig:unimodal_entropy}. The data provides strong empirical support for our methodology through three key observations. First, entropy consistently increases with higher design tiers ($d_v, d_t$), proving it aligns with our intended difficulty structure. This trend is especially clear in the vision modality, where for instance, the LLaVA-v1.6-7B model's entropy climbs steadily from approximately 0.25 at the lowest difficulty tier to over 1.5 at the highest. Second, the entropy values for both text and vision span a broad and comparable dynamic range from near-zero to over 1.75, which is crucial for creating conflict scenarios with diverse relative uncertainties. Third, and critically, the differences in entropy across models correspond to their known capabilities. The Qwen2.5-VL model, for example, consistently exhibits the lowest entropy, reflecting its strong performance, while we also observe expected scaling trends within model families, such as the LLaVA-v1.5-13B model showing generally lower visual uncertainty than its 7B counterpart.

\begin{takeaways}
\textbf{Conclusion:}
(1) We construct a novel dataset that allows for the systematic and independent control of reasoning difficulty across visual and textual modalities.
(2) Output token entropy is a robust and reliable proxy for a model's perceived unimodal uncertainty, establishing it as a sound foundation for our analysis.
\end{takeaways}

\begin{figure}[htbp]
  \centering
  \begin{subfigure}[t]{0.49\textwidth}
    \centering
    \includegraphics[width=\linewidth]{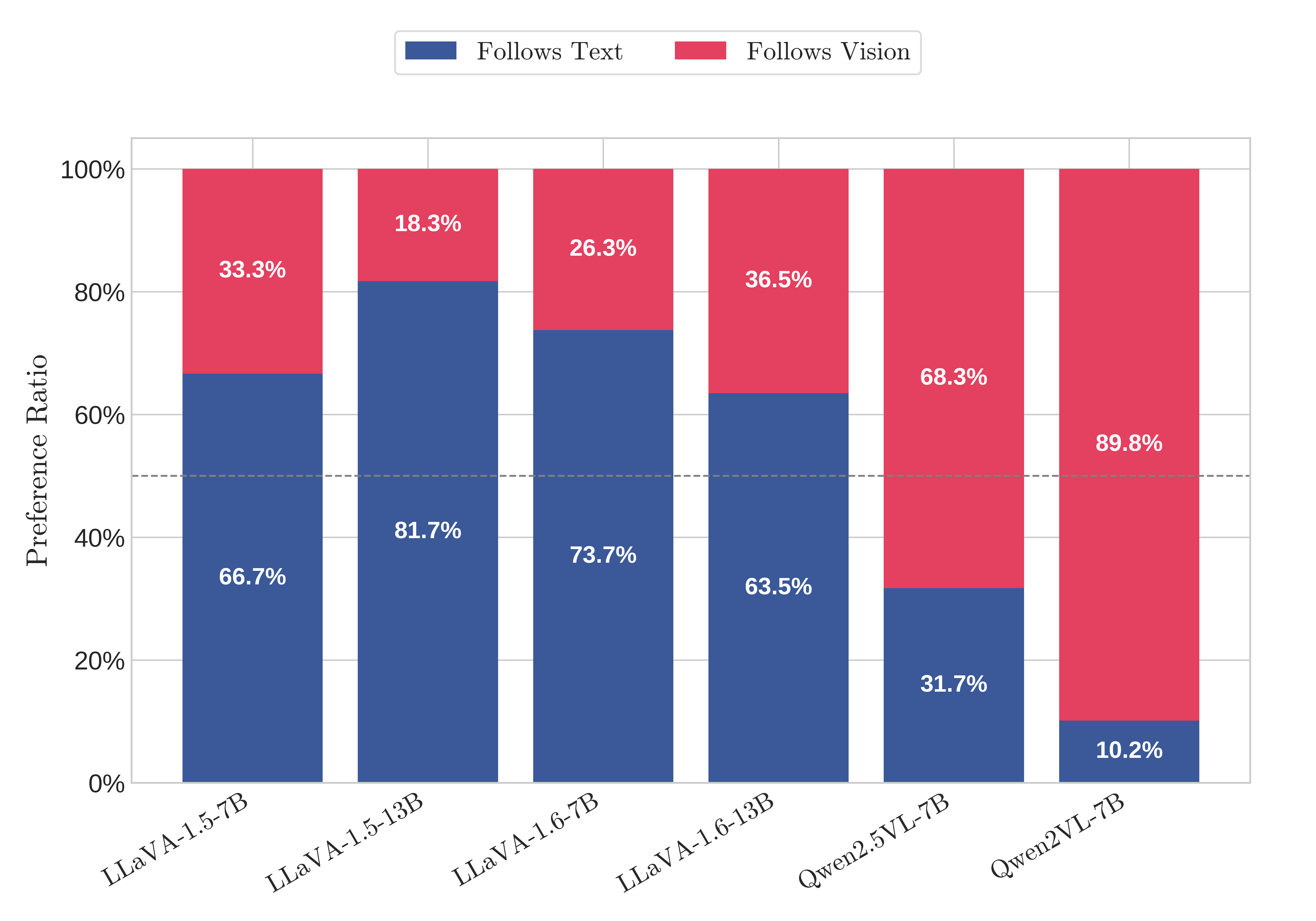}
    \caption{Overall macro-level performance}
    \label{fig:macro_a}
  \end{subfigure}
  \hfill
  \begin{subfigure}[t]{0.49\textwidth}
    \centering
    \includegraphics[width=\linewidth]{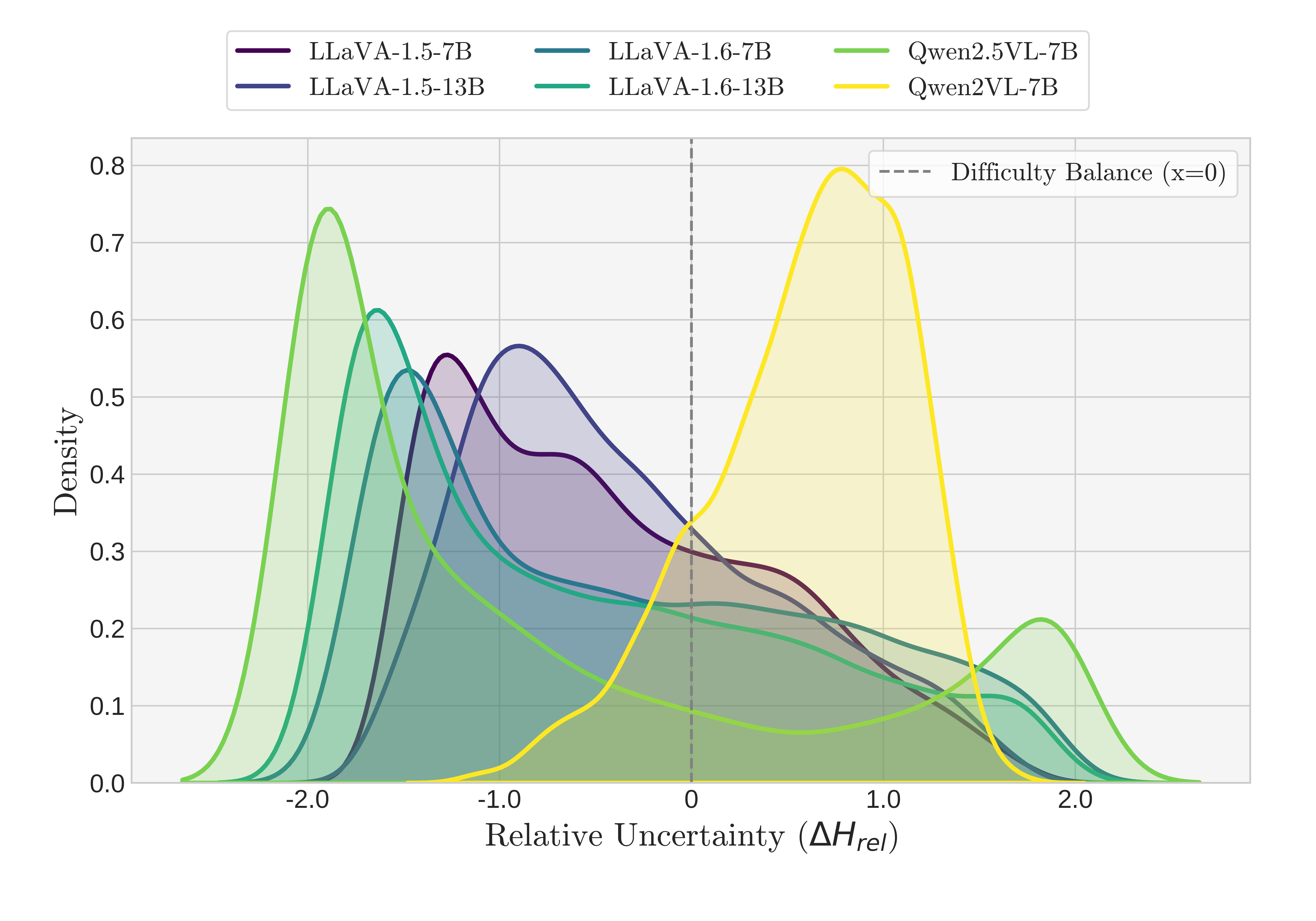}
    \caption{Relative uncertainty distribution}
    \label{fig:macro_b}
  \end{subfigure}
  \caption{Macro-level modality-following ratios and relative uncertainty distributions of model performance on the dataset.}
  \label{fig:macro_performance}
\end{figure}
\section{Modality Following is Shaped by Relative Uncertainty}
\label{sec:monotonic}
\paragraph{Contradictory Behaviors at the Macro Level.}
As a first step, we evaluate the modality-following behavior of six MLLMs using the text-following ratio (TFR), as defined in Section~\ref{sec:preliminary}. The types of MLLMs covers LLaVA1.5 Family \citep{llava1.5}, LLaVA1.6 Family \citep{li2024llava} and QwenVL family \citep{qwen2vl,qwen2.5,Qwen2.5-VL}. For this analysis, we focus on the subset of instances where the model answers correctly in both the vision-only and text-only settings. Figure~\ref{fig:macro_a} reveals stark, seemingly arbitrary differences between model families. The LLaVA series consistently exhibits a high TFR, appearing strongly text-following. In contrast, the Qwen-VL series is more vision-following. This raises a puzzle: \emph{why do models exhibit such divergent and seemingly fixed preferences when evaluated on the same dataset?}

\paragraph{A Finer Lens: Relative Unimodal Uncertainty.}
The core flaw in macro-level statistics like TFR is that they ignore the model's case-by-case reasoning confidence. To capture this, we introduce \textbf{relative unimodal uncertainty} ($\Delta H_{\mathrm{rel}}$). For a given conflicting input $x=(I, T, Q)$, we first decouple its components to measure the unimodal uncertainties. We calculate the text-only entropy, $H^{(t)}$, by providing only the text and the question $(T, Q)$ to the model. Similarly, we calculate the vision-only entropy, $H^{(v)}$, by providing only the image and the question $(I, Q)$. The relative uncertainty is the normalized difference between these two values:
\[
\Delta H_{\mathrm{rel}}(x) = \frac{2\big(H^{(t)}(x)-H^{(v)}(x)\big)}{H^{(t)}(x)+H^{(v)}(x)}.
\]
Here, $H^{(t)}(x)$ and $H^{(v)}(x)$ refer to the unimodal entropies derived from the components of the multimodal input $x$. This metric, $\Delta H_{\mathrm{rel}}$, thus quantifies the model's perceived confidence gap for each specific input. It is a direct manifestation of the model's \textbf{unimodal capabilities}, shaped by its architecture and training data. A negative value indicates the model is more confident in the text, while a positive value means it is more confident in the vision. When we plot the distribution of $\Delta H_{\mathrm{rel}}$ for the correctly solved cases (Figure~\ref{fig:macro_b}), a new puzzle emerges. Despite their different macro-level behaviors, most models face a similar distribution skewed towards negative values, meaning the dataset is, on average, easier for them to process through text. This deepens the mystery: \emph{if the underlying difficulty distribution is similar for most models, why are their final choices so different?}

\begin{figure}[!t]
  \centering
  \begin{subfigure}[t]{0.49\textwidth}
    \centering
    \includegraphics[width=\linewidth]{images/llava_comparison_unmodified.png}
    \caption{TRP decreases monotonically with relative uncertainty ($\Delta H_{rel}$). Each model's unique balance point (where its curve crosses the 0.5 probability line) quantifies its inherent preference.}
    \label{fig:curve_a}
  \end{subfigure}
  \hfill
  \begin{subfigure}[t]{0.49\textwidth}
    \centering
    \includegraphics[width=\linewidth]{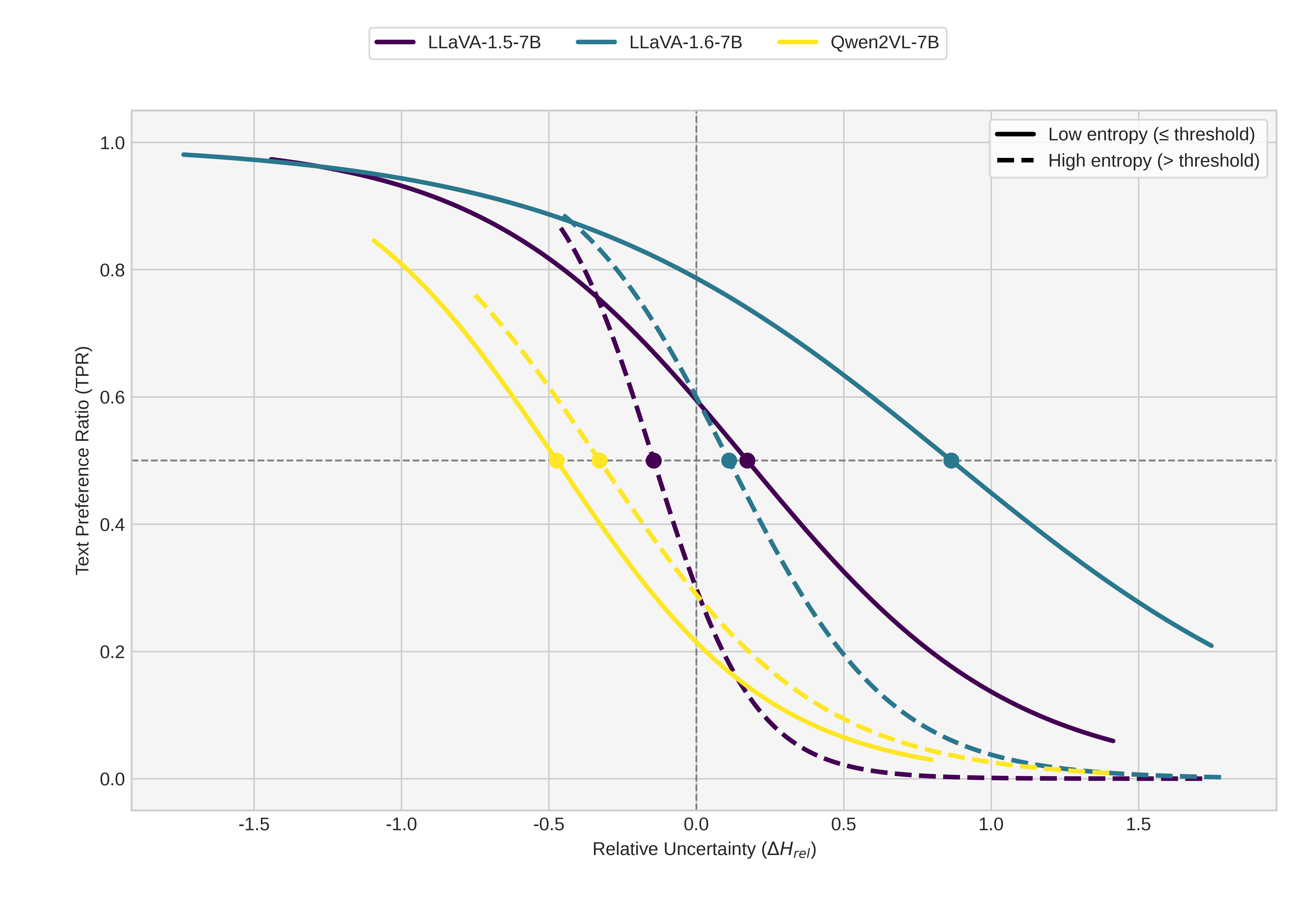}
    \caption{The monotonic law remains robust when data is split into low-entropy (solid lines) and high-entropy (dashed lines) subsets.}
    \label{fig:curve_b}
  \end{subfigure}
  \caption{The relationship between relative unimodal uncertainty ($\Delta H_{\text{rel}}$, x-axis) and the probability of following the text modality (Text Preference Ratio, y-axis) for various models. }
  \label{fig:curve}
\end{figure}

\paragraph{A Unified Monotonic Law.}
The answer emerges when we shift our perspective from aggregate statistics to the dynamic relationship between uncertainty and choice. By plotting the probability of a model following the text modality against the corresponding $\Delta H_{\mathrm{rel}}$ for each case, the apparent chaos resolves into a single, unified pattern, as shown in Figure~\ref{fig:curve_a}. For all six models, regardless of architecture or scale, the curve shows a smooth, \textbf{monotonic decrease}. In other words, as text becomes harder relative to vision (i.e., as $\Delta H_{\mathrm{rel}}$ increases), the probability that the model follows the text steadily and predictably decreases. This discovery directly confirms our central hypothesis from the Introduction: modality following is not a fixed trait but a dynamic behavior governed by relative reasoning uncertainty.

\paragraph{Quantifying Inherent Preference via the Balance Point.}
While all models obey this monotonic law, their curves are positioned differently along the axis. This leads to our second key insight. We define the \textbf{balance point} as the $\Delta H_{\mathrm{rel}}$ value at which the model is equally likely to follow either modality (a 50\% text-following probability). This balance point provides a principled, quantitative measure of the model's \textbf{inherent modality preference}—the concept we introduced in the Introduction as the model's intrinsic leaning when reasoning difficulty is equalized. A balance point below zero indicates an inherent \emph{vision preference} (as text must be significantly easier to be treated as equal), while a point above zero indicates an inherent \emph{text preference}. This finally allows us to disentangle a model's fluid, in-the-moment decision-making from its stable, underlying biases.

\paragraph{Reconciling Macro-Level Contradictions.}
Our framework, which separates unimodal capability (reflected in the $\Delta H_{\mathrm{rel}}$ distribution) from inherent preference (the balance point), can now fully explain the apparent contradictions from our initial macro-level analysis. Consider Qwen2-VL, which appears more vision-following than Qwen2.5-VL based on its VFR. Our analysis reveals this is largely a dataset artifact. Qwen2-VL's stronger visual capabilities on this specific dataset mean that more data points simply fall into the "vision-is-easier" (positive $\Delta H_{\mathrm{rel}}$) region, mechanically inflating its vision-following stats. However, Qwen2.5-VL has a balance point further to the left (more negative), revealing a \emph{stronger inherent vision preference}, as it continues to trust vision even when text is substantially easier. Similarly, the difference between LLaVA and Qwen models is not just about capability. While both face a dataset where text is often easier, Qwen models possess a clear inherent vision preference (negative balance point), whereas LLaVA models have a neutral or text-leaning preference (balance point near or above zero). It is this crucial difference in their \emph{inherent preference} that drives their divergent behaviors, a nuance entirely missed by macro-level metrics.

\paragraph{Robustness and Generality.}
To test the generality of our findings, we verified that the monotonic law remains stable across different conditions. We split the data into high- and low-entropy subsets (based on the median total entropy). As shown in Figure~\ref{fig:curve_b}, both subsets preserve the same monotonic decline, with only minor shifts in balance points: 
in high-entropy cases, the balance point moves closer to the center, consistent with the intuition that an already uncertain modality is more easily swayed by relative difficulty in the other.  Furthermore, evaluations on additional benchmarks, including our attribute-recognition dataset and tasks from the MC$^2$ benchmark, consistently revealed the same monotonic pattern (see Appendix~\ref{app:curve_of_all}). This confirms that the relationship between relative uncertainty and modality following is a robust and general principle.

\begin{takeaways}
\textbf{Takeaways:}
(1) Seemingly arbitrary macro-level following behaviors can be explained by a single, unified principle: the probability of following a modality monotonically decreases as its relative reasoning uncertainty increases.
(2) A model's \textbf{inherent preference} can be quantified as the ``balance point" on the relative uncertainty axis, separating it from the confounding effects of unimodal capability and dataset distribution.
(3) Traditional macro-level metrics (like TFR/VFR) are misleading because they conflate these two distinct factors: the model's capabilities and its inherent preference. Our framework successfully disentangles them.
\end{takeaways}
\begin{figure}[!t]
  \centering
    \includegraphics[width=0.75\linewidth]{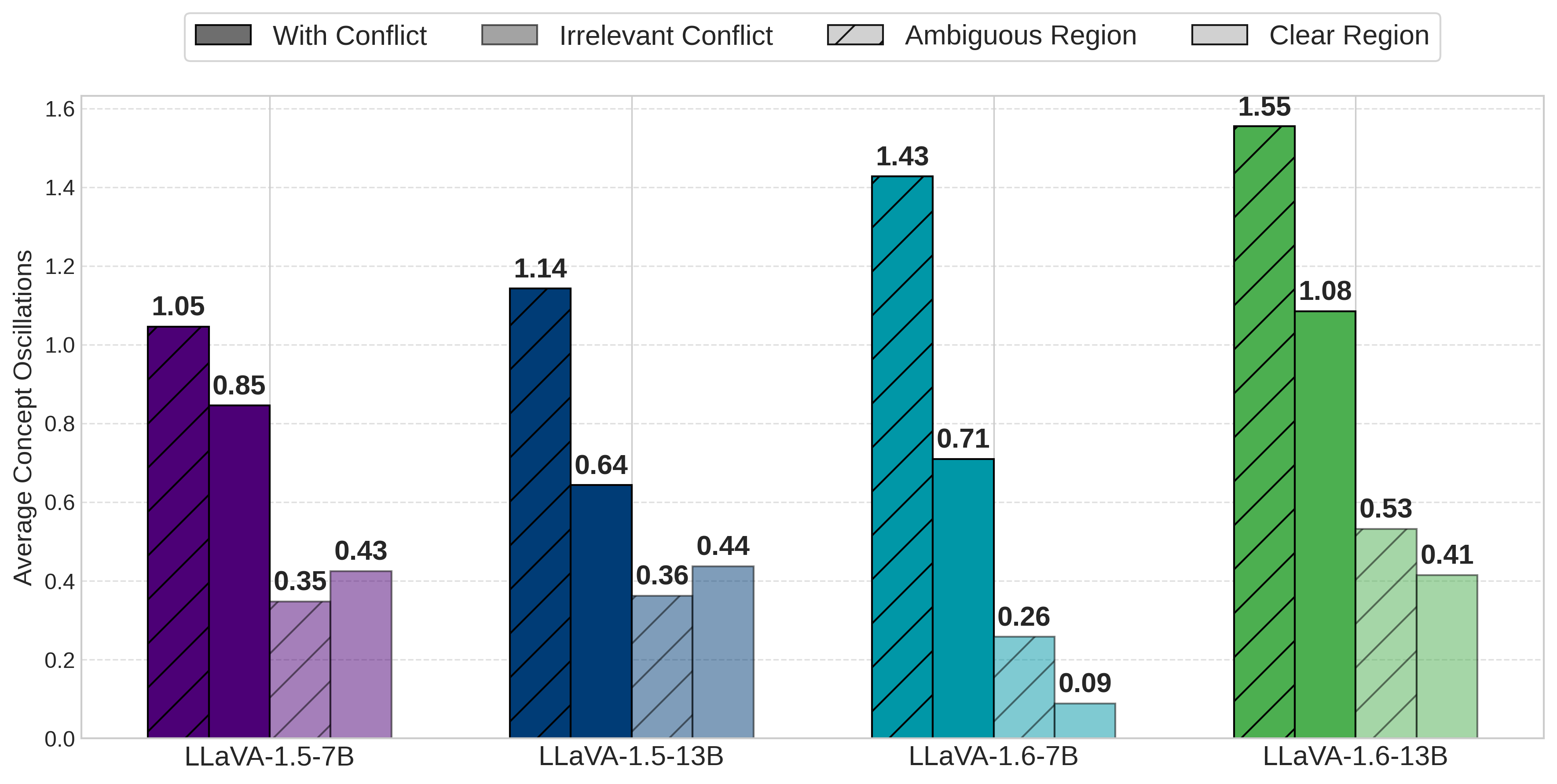}
    \caption{A comparison of the average number of concept oscillations for different models. Across all models, the number of oscillations is significantly higher in the ambiguous region (patterned bars) than in the clear region (solid bars).}
  \label{fig: oscillation}
\end{figure}
\section{The Internal Mechanism: Oscillation in the Face of Ambiguity}

Having established a robust behavioral law that modality following is a dynamic function of relative uncertainty, we now turn to the underlying mechanism. \emph{Why} does a model hesitate and produce averaged following behavior when the relative uncertainty is close to its inherent balance point? In this section, we peer inside the model's layer-by-layer reasoning process to reveal the internal dynamics of its decision-making. Our analysis demonstrates that the model's external hesitation is a direct consequence of internal \textbf{oscillations} between the conflicting choices.

\paragraph{Probing Layer-wise Predictions in Ambiguous vs. Clear Regions.}
To quantify the model's internal decision process, we conducted two analyses. First, we defined distinct reasoning scenarios. A case is in the \emph{ambiguous region} if its relative uncertainty $\Delta H_{\mathrm{rel}}$ is within a 0.5 radius of the model's balance point; otherwise, it is in the \emph{clear region}, where one modality is significantly easier. Second, we tracked the model's top-1 prediction for the answer token at each layer using a \emph{LogitLens}-style technique \citep{LogitLens2020,zhang2024locate}. Finally, to quantify this internal struggle, we define and count the number of \textbf{oscillations}. An oscillation is counted whenever the model's layer-wise top-1 prediction switches from a vision-supported answer to a text-supported answer, or vice-versa, regardless of any intermittent predictions of irrelevant tokens. For instance, a sequence of layer-wise predictions like `vision $\to$ irrelevant $\to$ text` counts as a single oscillation. This robust definition captures the number of times the model vacillates between the two primary conflicting concepts.
To ensure our analysis captures true semantic conflict, we also designed a control group with \textbf{irrelevant conflict}, where the text describes a different object with a conflicting attribute (e.g., for a red square, the text becomes ``The triangle is blue''). This maintains sentence structure while removing the direct conflict about the target object.

The results shown in Figure~\ref{fig: oscillation} reveal that the irrelevant conflict group consistently shows a very low number of oscillations (e.g., 0.35 for LLaVA-1.5-7B), confirming that the struggle is not due to mere sentence structure but to the semantic contradiction itself. More importantly, across all models, the ambiguous region with conflict exhibits significantly more oscillations than the clear region. For LLaVA-1.6-7B, the oscillation count in the ambiguous region (1.43) is nearly double that of the clear region (0.71), providing strong statistical evidence that models vacillate when faced with choices of similar perceived difficulty.

\begin{figure}[!t]
  \centering
  \begin{subfigure}[t]{0.49\textwidth}
    \centering
    \includegraphics[width=\linewidth]{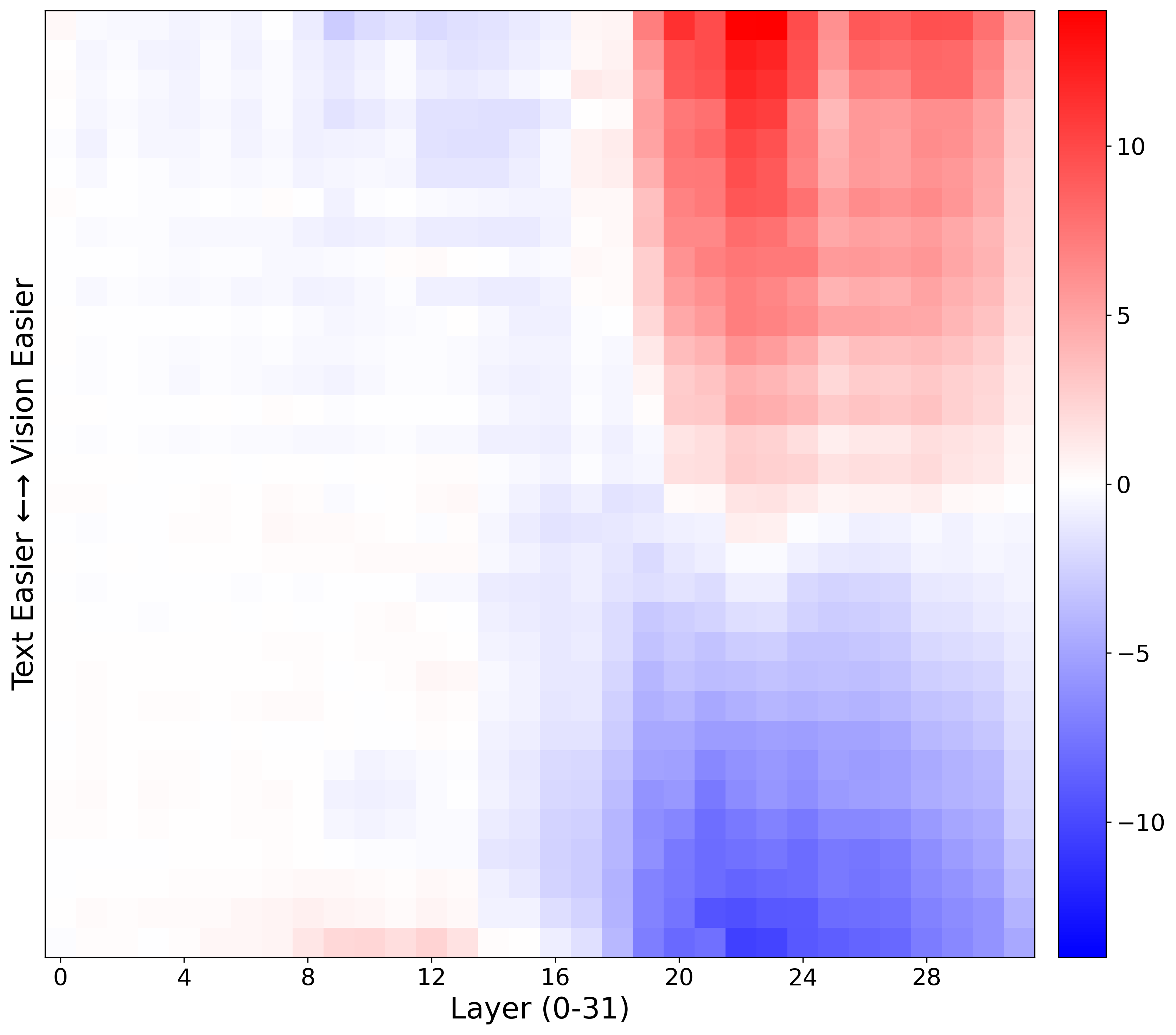}
    \caption{Logit Difference Heatmap Across Model Layers and Relative Uncertainty.}
    \label{fig:heatmap_logitdiff}
  \end{subfigure}
  \hfill
  \begin{subfigure}[t]{0.49\textwidth}
    \centering
    \includegraphics[width=\linewidth]{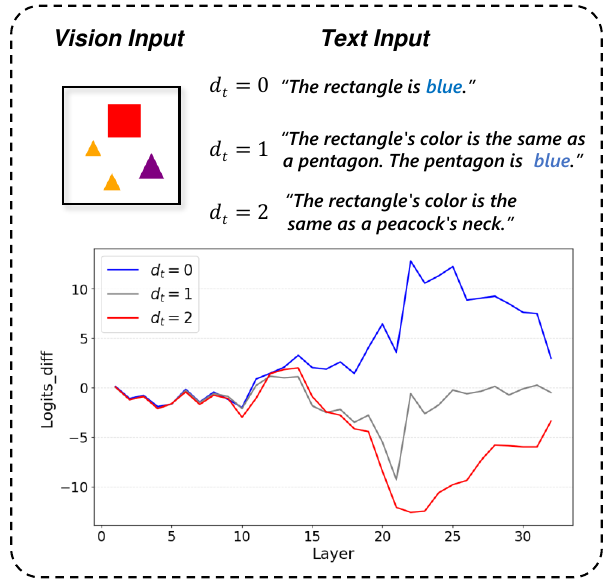}
    \caption{Case Study: Impact of Text Uncertainty on Layer-wise Confidence Dynamics. }
    \label{fig:logitdiff_case}
  \end{subfigure}
  \caption{
Visualization of the Model's Internal Decision-Making Dynamics. In these visualizations, the x-axis represents the model's layers. The y-axis is the logit difference, calculated as the logits of the text answer minus the logits of the vision answer ($\text{logit}(Y_t) - \text{logit}(Y_v)$).}
  \label{fig:curve}
\end{figure}
\paragraph{Visualizing Indecision with Logit Difference Heatmaps.}
To further investigate this internal struggle, we examine the difference in logits between the text-supported answer and the vision-supported answer across all layers. Figure~\ref{fig:heatmap_logitdiff} presents a heatmap of this logit difference. The x-axis represents the model's layers, and the y-axis represents the relative uncertainty $\Delta H_{\mathrm{rel}}$. The heatmap provides two key insights. First, near the center of the y-axis (the ambiguous region), the logit difference remains close to zero for many layers (indicated by the white color), meaning the model is highly uncertain. This numerical indecision is the direct cause of the oscillations. Second, towards the extremes of the y-axis (the clear regions), the color deepens to solid red or blue in the early-to-mid layers. This shows that when one modality is clearly easier, the model quickly and confidently commits to its corresponding answer, leading to stable processing.

\paragraph{A Case Study: The Dynamics of Conflict in a Single Image.}
Finally, we return to a concrete example to demonstrate our findings in action. Figure~\ref{fig:logitdiff_case} plots the layer-wise logit difference for a single visual input paired with three text prompts of increasing reasoning difficulty ($d_t=0, 1, 2$). By manipulating $d_t$, we effectively place the model into three distinct regions on the relative uncertainty spectrum, revealing its dramatically different internal states. The easy text ($d_t=0$) places the model in the \emph{text-clear region}, and its trajectory (the \emph{blue line}) shows a rapid, stable commitment to the text modality. Conversely, the hard text ($d_t=2$) pushes the case into the \emph{vision-clear region}, where the \emph{red line} decisively commits to vision. Most importantly, the intermediate difficulty text ($d_t=1$) creates an \emph{ambiguous region} case; its trajectory (the \emph{gray line}) visualizes the internal hesitation and oscillation by hovering near the zero-line decision boundary. This single example encapsulates our central thesis: controllable input difficulty ($d_t$) shapes relative uncertainty, which in turn determines the model's internal state and its final, observable choice.
\section{Related Work}

\paragraph{Processing and Characterizing Conflicting Information.}
A significant body of research has focused on characterizing how Multimodal Large Language Models (MLLMs) behave when faced with conflicting inputs. Various benchmarks have been developed to probe this phenomenon, revealing a complex and often inconsistent landscape of modality preferences. A frequently reported observation is that many models exhibit a ``blind faith'' in text, systematically ignoring visual information \cite{deng2025wordsvisionvisionlanguagemodels}. However, this tendency is not universal, as other studies demonstrate that preferences can vary significantly across different models and scenarios \cite{zhang2025evaluatingsteeringmodalitypreferences, liu2024insightsightexploringvisionknowledge}. Further work with benchmarks like MMIR has focused on the model's ability to detect and reason about such inconsistencies \citep{yan2025multimodalinconsistencyreasoningmmir}. The lack of a consistent principle to explain these varied and often contradictory macro-level observations is a key motivation for our work. Our primary contribution is to move beyond dataset-level statistics by proposing a unifying framework. We explain this apparent variability as an emergent property of two core factors: case-specific \textbf{relative reasoning uncertainty} and a model's stable \textbf{inherent preference}.

\paragraph{Explaining and Interpreting Conflict Resolution.}
Another line of research seeks to explain the underlying causes of modality preference. Some studies focus on external factors that can steer a model's behavior, such as the order of inputs \citep{deng2025wordsvisionvisionlanguagemodels} or the use of instructional prompts. Others delve deeper, attributing the behavior to internal factors like inconsistencies within the model's learned knowledge representations \cite{zhu2024unraveling, golovanevsky2025pixels}. A third approach uses attribution methods, such as those based on Shapley values, to quantify the relative influence of each modality on the final decision \citep{alishahi2019analyzing,parcalabescu2022mm,parcalabescu2024vision}. While these approaches identify potential causes and influencing factors, they do not fully reveal the dynamic, layer-by-layer computational process through which a model resolves ambiguity. Motivated by this gap, our work provides this missing mechanistic link. We introduce the concept of internal ``oscillations'' as direct, observable evidence of the conflict resolution process, demonstrating how our high-level framework is physically manifested in the model's computational dynamics and explains \emph{why} models hesitate under uncertainty.

\section{Conclusion}

Prior investigations of modality following have typically relied on coarse dataset-level statistics, often ignoring how differences in unimodal uncertainty shape aggregate outcomes. Without explicitly accounting for or aligning uncertainty across modalities, such analyses risk conflating a model’s capabilities with its underlying biases. We reframed modality following in MLLMs as a dynamic process shaped jointly by relative reasoning uncertainty and inherent modality preference. Across models and datasets, we uncovered a robust law: the likelihood of following a modality monotonically decreases as its relative uncertainty grows, with the balance point offering a principled measure of inherent preference. Probing layer-wise predictions further revealed that in ambiguous regions near this balance point, models exhibit strong oscillations between modalities, directly explaining their external hesitation. This framework thus disentangles capability from preference and provides a clearer lens for understanding and improving MLLM decision dynamics.




\bibliography{iclr2025_conference}
\bibliographystyle{iclr2025_conference}

\newpage
\appendix
\section{The Use of Large Language Models (LLMs)}

During the preparation of this paper, we used large language models (LLMs) solely as general-purpose writing assistants. Specifically, LLMs were employed to help refine the clarity, grammar, and readability of our drafts, as well as to suggest alternative phrasings in English. Importantly, all conceptual contributions including the design of research questions, development of methods, execution of experiments, and interpretation of results were conceived and carried out entirely by the authors. The authors carefully reviewed and edited all text suggested by LLMs to ensure accuracy and originality, and we take full responsibility for the final content of the paper.

\section{Information Conflict Dataset Generation Detials}

To investigate the external performance and internal mechanisms of multimodal models when dealing with conflicts between image and text information, we constructed two datasets. The first is \textbf{Color Recognition Dataset}, which requires the model to identify the color of geometric shapes placed on a white canvas. The second is \textbf{Attribution Recognition Dataset}, adapted and filtered from the CLEVR\citep{johnson2017clevr} dataset, whose task is to identify the material and shape of three-dimensional objects. Both datasets contain multiple task groups. Each group provides images with increasing visual complexity and text descriptions that contradict the image information while exhibiting increasing textual reasoning complexity. By systematically controlling the visual perception complexity ($d_v$) and the textual reasoning complexity ($d_t$), this design constructs conflict scenarios with diverse visual-textual difficulty combinations in a systematic manner.

\subsection{Dataset Overview}
\label{app:dataset_stats}
The Color Recognition Dataset consists of 400 groups, each containing 14 images and questions with 3 different types of conflict descriptions. Images with difficulty levels 0–4 are 800×600 pixels, while those with levels 5–13 are 224×224 pixels. The text is divided into three different types, with an average length of 22.7 words. In each group, the same image\_answer color can be derived from any image information, while the same text\_answer color which is different from the image\_answer, can be obtained from any conflict description in the text. The distribution of image\_answer and text\_answer is as follows:
\begin{itemize}
    \item \textbf{Image\_answer Colors:} Red(67), Yellow(67), Blue(67), Green(66), Purple(66), Orange(67)
    \item \textbf{Text\_answer Colors:} Red(67), Yellow(66), Blue(67), Green(67), Purple(66), Orange(67)
\end{itemize}

The Shape subset and the Material subset of the Attribution Recognition Dataset each contain 300 groups. Each group includes 4 images and questions with 3 different types of conflict descriptions. All images are 480×320 pixels, while the text is divided into five different types, with an average length of 30.0 words. In each group, the same image\_answer attribute can be derived from any image information, while the same text\_answer attribute which is different from the image\_answer can be obtained from any conflict description in the text. The distribution of image\_answer and text\_answer is as follows:
\begin{itemize}
    \item \textbf{Image\_answer Shapes:} Sphere(108), Cube(100), Cylinder(92)
    \item \textbf{Text\_answer Shapes:} Sphere(100), Cube(92), Cylinder(108)
    \item \textbf{Image\_answer Materials:} Metal(160), Rubber(140)  
    \item \textbf{Text\_answer Materials:} Metal(140), Rubber(160) 
\end{itemize}

\subsection{Image Generation of Color Recognition Dataset}

For each set of 14 images with a progressive difficulty gradient in the Color Recognition Dataset, we used the Python PIL library for rendering. The following is the generation pipeline.

\begin{enumerate}
    \item \textbf{Initialization:} A \textbf{target shape} (e.g., Circle) is randomly selected.
    \item \textbf{Color Assignment:}
    \begin{itemize}
        \item \textbf{Visual Answer Color:} One color is randomly assigned to the target shape.
        \item \textbf{Textual Answer Color:} A different color is randomly selected as the conflicting textual statement.
    \end{itemize}
    \item \textbf{Distractor Generation:} Distractor shapes are randomly chosen from the set excluding the target shape. Their colors are randomly selected from the set excluding both the visual and textual answer colors.
    \item \textbf{Difficulty Tiers (\(d_v = 0\) to \(13\)):} Fourteen progressive difficulty levels are defined by target size, number of distractors and occlusion. Parameters are specified in Table\ref{tab:visual_difficulty_color}.
\end{enumerate}
\begin{table}[h!]
\centering
\caption{Visual Difficulty (\(d_v\)) Tiers Specification}
\label{tab:visual_difficulty_color}
\begin{tabular}{|c|c|c|c|l|}
\hline
\textbf{Difficulty(\(d_v\))} & \textbf{Target Size} & \textbf{\# Distractors} & \textbf{Occlusion Rule} \\ \hline
0 & 80-200 pixels & 0 & No occlusion \\ \hline
1 & 80-200 pixels & 1 & No occlusion \\ \hline
2 & 80-200 pixels & 2 & No occlusion \\ \hline
3 & 80-200 pixels & 3 & No occlusion \\ \hline
4 & 80-200 pixels & 4 & No occlusion \\ \hline
5 & 20\%-40\% of image & 7 & 50\% occlusion rate \\ \hline
6 & 20\%-40\% of image & 10 & 80\% occlusion rate \\ \hline
7 & 5\%-10\% of image & 7 & 50\% occlusion rate \\ \hline
8 & 5\%-10\% of image & 11 & 80\% occlusion rate \\ \hline
9 & 4\%-6\% of image & 20 & 30\% occlusion rate \\ \hline
10 & 4\%-6\% of image & 30 & 60\% occlusion rate \\ \hline
11 & 4\%-6\% of image & 40 & 50\% occlusion rate \\ \hline
12 & 4\%-6\% of image & 55 & 60\% occlusion rate \\ \hline
13 & 4\%-6\% of image & 70 & 70\% occlusion rate \\ \hline
\end{tabular}
\smallskip
\parbox{\linewidth}{
\scriptsize \textbf{Note 1:} "Occlusion rate" refers to the proportion of distractors that visually overlap the target. Different rates for odd/even tiers introduce finer-grained difficulty variation.}
\end{table}

\subsection{Image Selection of Attribution Recognition Dataset}
All images in the Attribution Recognition Dataset were curated from the CLEVR dataset, which contains objects defined by three geometric shapes (cube, sphere, cylinder), two materials (rubber, metal), and eight colors. For each target attribute corresponding to the subset, our selection procedure began by forming all possible attribute--color pairs via the Cartesian product. For each unique pair, we identified images from the CLEVR validation set containing \emph{exactly one} object matching that specific combination. The selected images were then assigned a difficulty level based on scene complexity, with a fixed number of images sampled per level to construct the final task groups.Table\ref{tab:visual_difficulty_3D} shows the various difficulty levels of the pictures.

\begin{table}[ht]
  \centering
  \caption{Difficulty levels for image selection}
  \label{tab:visual_difficulty_3D}
  \begin{tabular}{@{}lll@{}}
    \toprule
    Difficulty(\(d_v\)) & Number of objects in scene & Target object size \\
    \midrule
    0 & 3--4 objects   & large \\
    1 & 6--8 objects   & large \\
    2 & 6--8 objects   & small \\
    3 & $\geq$10 objects & small \\
    \bottomrule
  \end{tabular}
\end{table}

\subsection{Textual Modality Construction}
\label{app:dataset_robustness}
The conflict text issues between the Color Recognition Dataset and the Attribution Recognition Dataset share many similarities in terms of structure and pipeline construction. In both cases, we gradually increase the complexity of the textual modality by increasing the number of reasoning steps and converting explicit reasoning into implicit reasoning. The questions within the same group share a fixed \textbf{target\_shape} with the images of that group, inquire an \textbf{attribute} depending on the dataset they belong to, and utilize an identical \textbf{text\_answer} that contradicts the image information. Each textual problem follows the format of: [Conflict Description] + [Question] + [Command].

\begin{itemize}
  \item \textbf{Question:} \texttt{What \{attribute\} is the \{target\_shape\}?}
  \item \textbf{Command:} \texttt{Please use one word to answer this question.}
\end{itemize}

For each group, we generate 3 types of conflict description for Color Recognition Dataset and 4 for Attribution Recognition Dataset with increasing difficulty. The Table\ref{tab:question_types} below lists each type and a concise description, where \textbf{A} denotes the target\_object, \textbf{T} denotes the text\_answer, \textbf{B/S1/S2} represent randomly selected objects absent from the image, \textbf{D} represents a real-world instance unambiguously possessing attribute T, and \textbf{Pos1/Pos2} denote a pair of opposite spatial relations Left and Right.

\begin{table}[ht]
  \centering
  \caption{Question types and descriptions (descriptions only)}
  \label{tab:question_types}
  \begin{tabular}{@{}>{\raggedright\arraybackslash}p{2cm}>{\raggedright\arraybackslash}p{3cm}>{\raggedright\arraybackslash}p{8cm}@{}}
    \toprule
    Difficulty(\(d_t\)) & Type & Description \\
    \midrule
    x & Original & No interference description. \\
    0 & Direct & The A is T. \\
    1 & Indirect\_simple & The A's \{attribute\} is the same as a B. The B is T. \\
    2 & Indirect & The A's \{attribute\} is the same as a D. \\
    3 & Space(Attribution Recognition Dataset only) & There is a T S1, on the {Pos1} of the S1 is a S2. The A's \{attribute\} is the same as the object {Pos2} to the S2. \\
    \bottomrule
  \end{tabular}
\end{table}

\textbf{Robustness Processing:} To prevent models from solving tasks via superficial pattern matching, texts in Color Recognition Dataset for \(d_t \geq 0\) were paraphrased using Qwen-Plus\citep{alibaba2025qwenplus}. This process preserved core semantics, reasoning structure, and key information tokens while varying sentence structure, prepositional phrases, and lexical choices.

\textbf{Control Group Setup:}For ablation studies, two types of control data were constructed:
\begin{itemize}
    \item \textbf{Text-Irrelevant:} The target shape `A` in conflict description only is replaced with a randomly chosen \textbf{non-target shape} (e.g., if target is `circle`, replace with `triangle` or `rectangle`).
    \item \textbf{Image-Irrelevant:} The target shape `A` in the entire text is replaced with a shape \textbf{never present} in the images (`star`, `cone`, `frustum`), maintaining the correspondence between the question and the text description while severing the connection with the image.
\end{itemize}
    
\begin{tcolorbox}[colback=gray!5!white,colframe=black!75!white,title=Rewrite Questions Task]
====SYSTEM==== \\
You are a conservative paraphrasing assistant specialized in subtle wording changes. 
Your goal is to rewrite a single question sentence while preserving *all* facts, *all* explicit instructions, and the exact multi-hop reasoning structure (number of inference steps and intermediate referents). Make only minor wording, grammar, punctuation, and token-count adjustments; do NOT add, remove, or transform factual content or the logical chain. \\

====USER====\\

Field type: \\
\{FIELD\_TYPE\} \\
Original question: \\
\{ORIGINAL\_QUESTION\} \\
Rewrite Instructions (STRICT): \\
1. Output exactly one rewritten question sentence (no explanation, no notes, no extra punctuation before/after). \\
2. Preserve *all* factual propositions and named referents. Do not add or remove facts. \\
3. Preserve the multi-hop reasoning structure: \\
   - If the original is a single-step (direct), keep it single-step. \\
   - If it is implicit multi-step (indirect), keep it implicit and do not make steps explicit. \\
   - If it is explicit multi-hop (indirect\_simple), keep the same explicit chain of premises and the same number of hops. \\
4. Preserve any explicit answering instruction exactly (e.g., "Please use one word to answer this question."). \\
5. Do not change the identity of entities (e.g., "hexagon", "pine tree", "circle") or the target attribute (e.g., "color"). \\
6. Only rewrite wording, punctuation, and sentence flow to be more natural or shorter, and optionally reduce/increase token count slightly. You can use near-synonyms with very high similarity. \\
7. Avoid introducing pronouns that obscure referents; keep clarity of which object each premise references. \\
8. If the original contains multiple sentences that together form the multi-hop chain, you may combine or split them only if you exactly preserve the same premises and hop order. \\
Output: the single rewritten question sentence (no extra text).
\end{tcolorbox}

\subsection{Illustrative Samples from the dataset}
\begin{figure}[htbp]
  \centering
  \begin{subfigure}[b]{0.48\textwidth}
    \centering
    \textbf{Group41 Difficulty0}\\[1ex]
    \includegraphics[width=\textwidth]{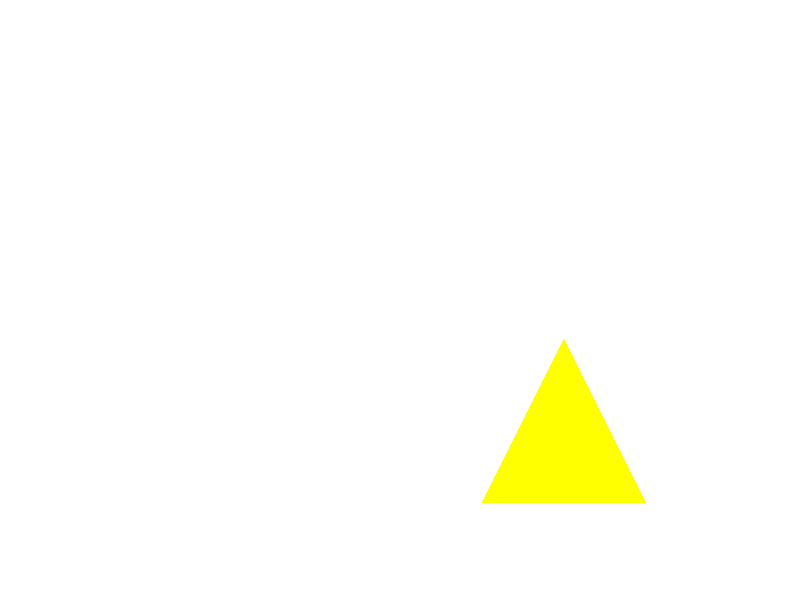}
    \caption*{
    \textbf{Original:}   \\ 
    \textbf{Question:} What color is the triangle? \\
    \textbf{Command:} Please use one word to answer this question. \\
      \textbf{Vision-based Answer:} \textit{Yellow}\\
      \textbf{Text-based Answer:} \textit{\textcolor[RGB]{202,12,22}{}}
    }
  \end{subfigure}
  \hfill
  \begin{subfigure}[b]{0.48\textwidth}
    \centering
    \textbf{Group41 Difficulty3}\\[1ex]
    \includegraphics[width=\textwidth]{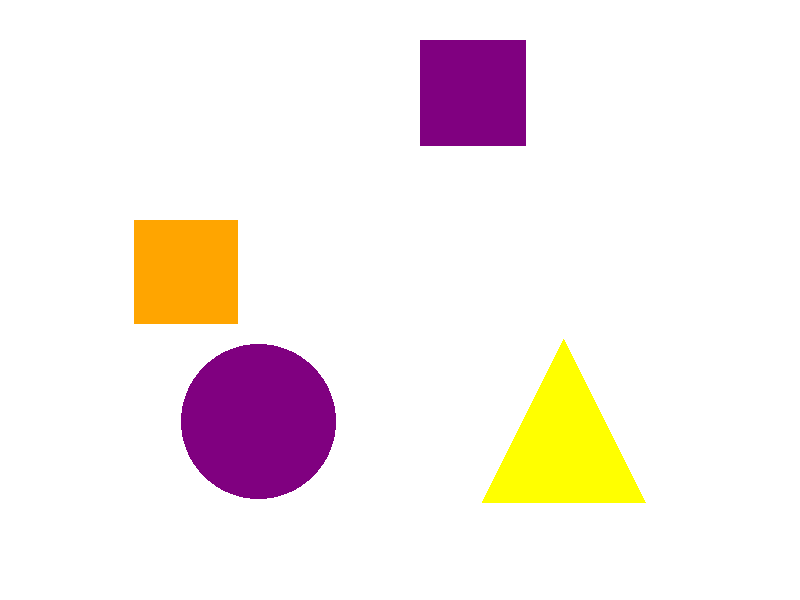}
    \caption*{
    \textbf{Direct:} \textcolor[RGB]{202,12,22}{The triangle is blue.}  \\ 
    \textbf{Question:} What color is the triangle? \\
    \textbf{Command:} Please use one word to answer this question. \\
      \textbf{Vision-based Answer:} \textit{Yellow}\\
      \textbf{Text-based Answer:} \textit{\textcolor[RGB]{202,12,22}{Blue}}
    }
  \end{subfigure}
  
  \vspace{2cm} 
  
  \begin{subfigure}[b]{0.48\textwidth}
    \centering
    \textbf{Group41 Difficulty6}\\[1ex]
    \includegraphics[width=\textwidth]{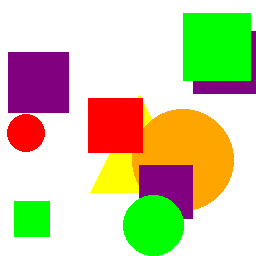}
    \caption*{
    \textbf{Indirect\_simple:}  \textcolor[RGB]{202,12,22}{The triangle's color is the same as a pentagon. The pentagon is blue.}  \\ 
    \textbf{Question:} What color is the triangle? \\
    \textbf{Command:} Please use one word to answer this question. \\
      \textbf{Vision-based Answer:} \textit{Yellow}\\
      \textbf{Text-based Answer:} \textit{\textcolor[RGB]{202,12,22}{Blue}}
    }
  \end{subfigure}
  \hfill
  \begin{subfigure}[b]{0.48\textwidth}
    \centering
    \textbf{Group41 Difficulty15}\\[1ex]
    \includegraphics[width=\textwidth]{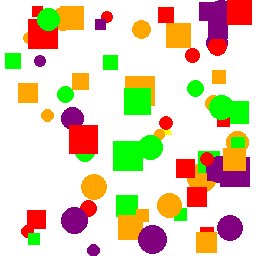}
    \caption*{
    \textbf{Indirect:} \textcolor[RGB]{202,12,22}{The triangle's color is the same as a mailbox in the US.}  \\ 
    \textbf{Question:} What color is the triangle? \\
    \textbf{Command:} Please use one word to answer this question. \\
      \textbf{Vision-based Answer:} \textit{Yellow}\\
      \textbf{Text-based Answer:} \textit{\textcolor[RGB]{202,12,22}{Blue}}
    }
  \end{subfigure}
\caption{A selection of image-text pairings from a group in the Color Recognition Dataset. The text highlighted in red indicates the descriptions and answers that conflict with the image information.}
\label{fig:ColorSampled}
\vspace{0.2cm}
\end{figure}

\begin{figure}[htbp]
  \centering
  \begin{subfigure}[b]{0.48\textwidth}
    \centering
    \textbf{Group193 Difficulty0}\\[1ex]
    \includegraphics[width=\textwidth]{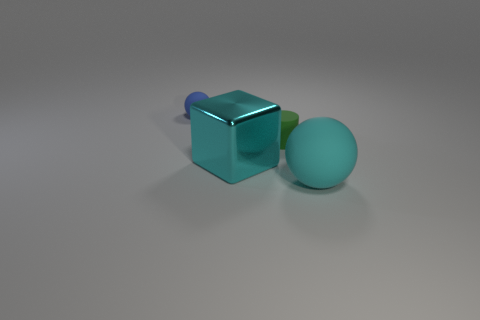}
    \caption*{
    \textbf{Direct:} \textcolor[RGB]{202,12,22}{The cyan rubber object is a cylinder.}  \\ \\
    \textbf{Question:} What is the shape of the cyan rubber object? \\
    \textbf{Command:} Please answer with one word. \\
      \textbf{Vision-based Answer:} \textit{sphere}\\
      \textbf{Text-based Answer:} \textit{\textcolor[RGB]{202,12,22}{cylinder}}
    }
  \end{subfigure}
  \hfill
  \begin{subfigure}[b]{0.48\textwidth}
    \centering
    \textbf{Group193 Difficulty2}\\[1ex]
    \includegraphics[width=\textwidth]{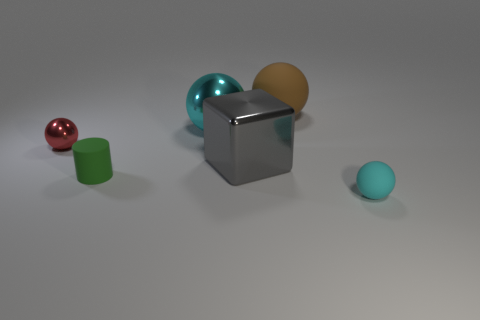}
    \caption*{
    \textbf{Indirect:} \textcolor[RGB]{202,12,22}{The cyan rubber object's shape is the same as a log.}  \\ 
    \textbf{Question:} What is the shape of the cyan rubber object? \\
    \textbf{Command:} Please answer with one word. \\
      \textbf{Vision-based Answer:} \textit{sphere}\\
      \textbf{Text-based Answer:} \textit{\textcolor[RGB]{202,12,22}{cylinder}}
    }
  \end{subfigure}
\caption{A selection of image-text pairings from a group in the Shape subset of the Attribution Recognition Dataset. The text highlighted in red indicates the descriptions and answers that conflict with the image information.}
\label{fig:ShapeSampled}
\vspace{0.2cm}
\end{figure}

\begin{figure}[htbp]
  \centering
  \begin{subfigure}[b]{0.48\textwidth}
    \centering
    \textbf{Group79 Difficulty1}\\[1ex]
    \includegraphics[width=\textwidth]{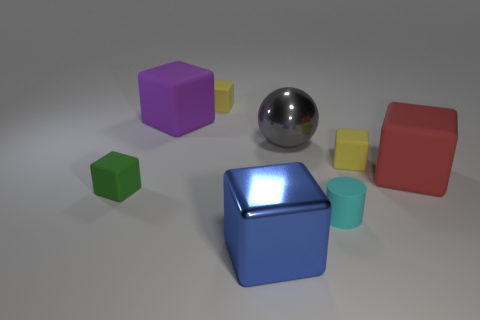}
    \caption*{
    \textbf{Indirect\_simple:} \textcolor[RGB]{202,12,22}{The Frustum is rubber, blue cube's material is the same as the Frustum.}  \\ \\
    \parbox{\linewidth}{\textbf{Question:} What is the material of the blue cube?} \\
    \textbf{Command:} Please use one word to answer this question. \\
      \textbf{Vision-based Answer:} \textit{metal}\\
      \textbf{Text-based Answer:} \textit{\textcolor[RGB]{202,12,22}{rubber}}
    }
  \end{subfigure}
  \hfill
  \begin{subfigure}[b]{0.48\textwidth}
    \centering
    \textbf{Group79 Difficulty3}\\[1ex]
    \includegraphics[width=\textwidth]{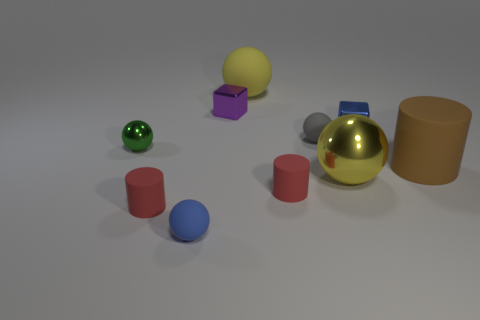}
    \caption*{
    \textbf{Space:} \textcolor[RGB]{202,12,22}{There is a rubber cone, the right of the cone is a wood frustum. The blue cube's material is the same as the object left to the wood frustum.}  \\ 
    \textbf{Question:} What is the material of the blue cube? \\
    \textbf{Command:} Please use one word to answer this question. \\
      \textbf{Vision-based Answer:} \textit{metal}\\
      \textbf{Text-based Answer:} \textit{\textcolor[RGB]{202,12,22}{rubber}}
    }
  \end{subfigure}
\caption{A selection of image-text pairings from a group in the Material subset of the Attribution Recognition Dataset. The text highlighted in red indicates the descriptions and answers that conflict with the image information.}
\label{fig:MaterialSampled}
\vspace{0.2cm}
\end{figure}

To provide a more intuitive understanding of our image-text conflict dataset, we have sampled several image-question pairs from the Color Recognition Dataset and the Attribution Recognition Dataset subsets and presented them in Figure\ref{fig:ColorSampled}, Figure\ref{fig:ShapeSampled} and Figure\ref{fig:MaterialSampled}.

\section{Curve of all remain datasets}
\label{app:curve_of_all}
\begin{figure}[htbp]
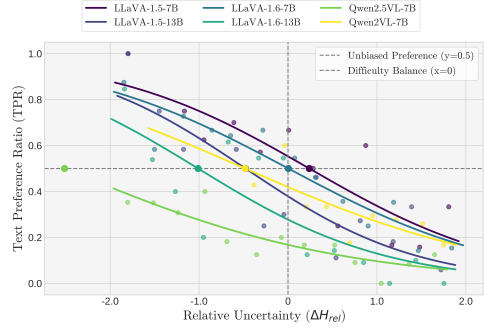
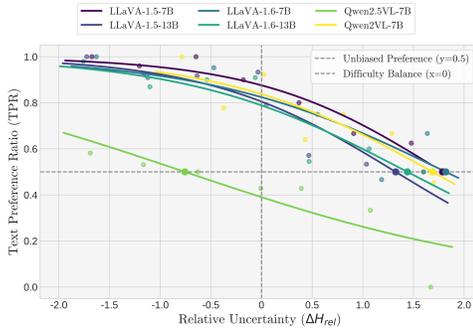
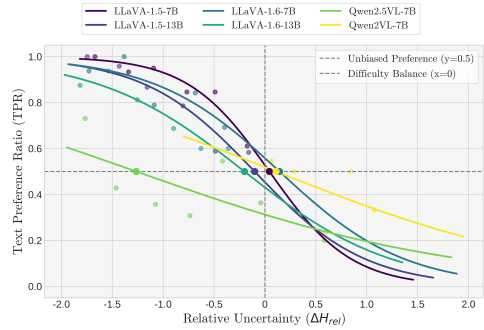
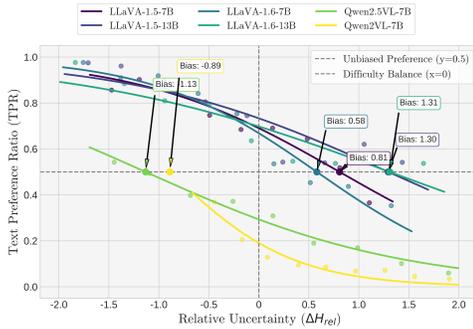
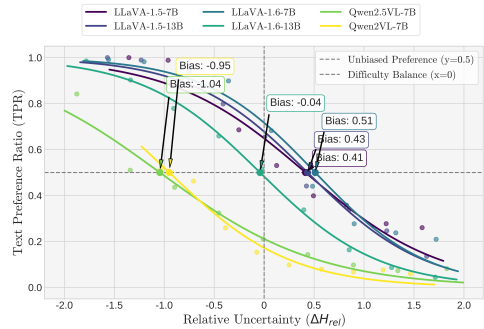

  \centering
  \begin{subfigure}[t]{0.49\textwidth}
    \centering
    \includegraphics[width=\linewidth]{images/real_dataset_curve/llava_comparison_color.png}
    \caption{Curve of Color Recognition Task in $MC^2$ Datasets.}
    \label{fig:curve_color}
  \end{subfigure}
  \hfill
  \begin{subfigure}[t]{0.49\textwidth}
    \centering
    \includegraphics[width=\linewidth]{images/real_dataset_curve/llava_compariso_object_recognition.png}
    \caption{Curve of Object Recognition Task in $MC^2$ Datasets.}
    \label{fig:curve_recognition}
  \end{subfigure}
  \hfill  
  \begin{subfigure}[t]{0.49\textwidth}
    \centering
    \includegraphics[width=\linewidth]{images/real_dataset_curve/llava_comparison_attribute.png}
    \caption{Curve of Attribution Recognition Task in $MC^2$ Datasets.}
    \label{fig:curve_attribute}
  \end{subfigure}
  \hfill
  \begin{subfigure}[t]{0.49\textwidth}
    \centering
    \includegraphics[width=\linewidth]{images/real_dataset_curve/llava_comparison_positional_reasoning.png}
    \caption{Curve of Position Reasoning Task in $MC^2$ Datasets.}
    \label{fig:curve_positional}
  \end{subfigure}
    \hfill
  \begin{subfigure}[t]{0.49\textwidth}
    \centering
    \includegraphics[width=\linewidth]{images/llava_comparison_ours_attribution.png}
    \caption{Curve of Attribution Recognition Task in Our Dataset.}
    \label{fig:curve_our_attribute}
  \end{subfigure}
    \hfill
  \begin{subfigure}[t]{0.49\textwidth}
    \centering
    \includegraphics[width=\linewidth]{images/llava_comparison_modified.png}
    \caption{Curve of Color Recognition Task in Our Dataset with prompts after rewriting.}
    \label{fig:curve_modified}
  \end{subfigure}
  \caption{Relative uncertainty versus text-following ratio (TFR) curves across multiple datasets, including \emph{Color Recognition}, \emph{Object Recognition}, \emph{Attribution Recognition}, and \emph{Position Reasoning} from the MC$^2$ benchmark, our CLEVR-derived \emph{Attribution Recognition} dataset introduced in Section~\ref{app:dataset_stats}, and the \emph{Color Recognition} dataset after prompt diversification with Qwen, which introduced in ~\ref{app:dataset_robustness}. Across all datasets and models, we consistently observe a monotonic decrease in TFR as relative uncertainty increases, confirming the robustness of the law. Meanwhile, the locations of the balance points vary significantly across datasets due to differences in textual and visual characteristics, which affect the resulting unimodal entropy distributions. These shifts in balance points reflect each model’s inherent preference toward the specific type of data.}
  \label{fig:real_curve}
\end{figure}


\end{document}